\documentclass[sigconf]{acmart}  

\makeatletter
\def\mdseries@tt{m}
\makeatother

\usepackage{booktabs} 

\setcopyright{rightsretained}

\newcommand{\Tref}[1]{Table~\ref{#1}}

\newcommand{\Fref}[1]{Fig.~\ref{#1}}
\newcommand{\Aref}[1]{Alg.~\ref{#1}}
\newcommand{\Sref}[1]{Sec.~\ref{#1}}
\DeclareMathOperator*{\argmin}{argmin}

\graphicspath{{./img/}}
\usepackage{subfig}
\usepackage[ruled, vlined, linesnumbered]{algorithm2e} 
\usepackage{multirow} 
\usepackage{wrapfig} 

\usepackage{enumitem}  
\newcommand{\ItemizeLeftMargin}{15pt}

\definecolor{DarkOliveGreen}{rgb}{0.33, 0.42, 0.18}
\newcommand{\codecommentcolor}{DarkOliveGreen} 
\newcommand{\codecomment}[1]{\textcolor{\codecommentcolor}{\texttt{#1}}}

\usepackage{color}

\usepackage[frozencache=true]{minted}

\begin{document}
\title{Reconfigurable Inverted Index}

\author{Yusuke Matsui}
\affiliation{%
	\institution{National Institute of Informatics}
}
\email{matsui@nii.ac.jp}

\author{Ryota Hinami}
\affiliation{%
	\institution{The University of Tokyo}
}
\email{hinami@nii.ac.jp}

\author{Shin'ichi Satoh}
\affiliation{%
	\institution{National Institute of Informatics}
}
\email{satoh@nii.ac.jp}

\renewcommand{\shortauthors}{Y. Matsui et al.}

\begin{abstract}
Existing approximate nearest neighbor search systems
suffer from two fundamental problems that are of practical importance but have not received sufficient attention from the research community. 
First, although existing systems perform well for the whole database,
it is difficult to run a search over a subset of the database.
Second, there has been no discussion concerning the performance decrement after
many items have been newly added to a system.
We develop a reconfigurable inverted index (Rii) to resolve these two issues.
Based on the standard IVFADC system, we design a data layout such that 
items are stored linearly. 
This enables us to efficiently run a subset search 
by switching the search method to a linear PQ scan if the size of a subset is small.
Owing to the linear layout, the data structure can be dynamically adjusted
after new items are added, maintaining the fast speed of the system.
Extensive comparisons show that Rii achieves a comparable
performance with state-of-the art systems such as Faiss.
\end{abstract}

%

\copyrightyear{2018}
\acmYear{2018}
\setcopyright{acmcopyright}
\acmConference[MM '18]{2018 ACM Multimedia Conference}{October 22--26, 2018}{Seoul, Republic of Korea}
\acmBooktitle{2018 ACM Multimedia Conference (MM '18), October 22--26, 2018, Seoul, Republic of Korea}
\acmPrice{15.00}
\acmDOI{10.1145/3240508.3240630}
\acmISBN{978-1-4503-5665-7/18/10}

\fancyhead{}

\ccsdesc[300]{Information systems~Nearest-neighbor search}
\ccsdesc[300]{Information systems~Search engine indexing}
\ccsdesc[300]{Information systems~Multimedia and multimodal retrieval}
\ccsdesc[300]{Computing methodologies~Visual content-based indexing and retrieval}

\keywords{Approximate nearest neighbor search; inverted index; product quantization; subset search; reconfigure}

\maketitle

\section{Introduction}
\label{sec:intro}
In recent years, the approximate nearest neighbor search (ANN) has received increasing attention
from various research communities~\cite{tomm_gudmundsson2016}. 
Typical ANN systems operate in two stages.
In the offline phase, database vectors are stored in the ANN system.
These vectors may be converted to other forms, such as compact codes, for fast searching and efficient memory usage.
In the online querying phase, the system receives a query vector.
Similar items to the query are retrieved from the stored database vectors.
Their identifiers (and optionally their distances to the query) are then returned.
To handle large datasets, this search should be not only fast and accurate, but also memory efficient.

\begin{figure}
	\begin{center}
		\subfloat[Subset search]{\includegraphics[width=1.0\linewidth]{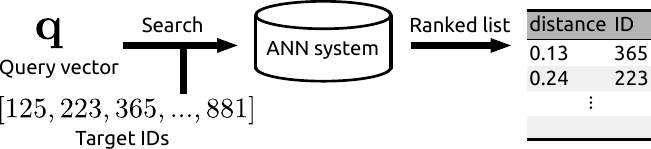}
			\label{fig:teaser1}} \\
		\subfloat[Performance degadation via data addition]{\includegraphics[width=1.0\linewidth]{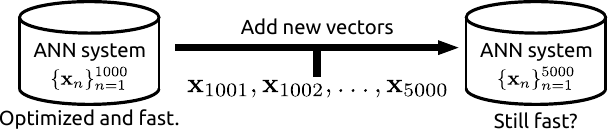}
			\label{fig:teaser2}}
	\end{center}
	\vspace{-2mm}
	\caption{
		The two problems tackled in this paper.
		(a) The search is operated for a subset of a database, which is specified by the target identifiers.
		The search result (ranked list) should contain the specified items only.
		(b) Given a fast (optimized) ANN system, new vectors are added. Is the updated ANN system still fast?}
	\label{fig:teaser}
	\vspace{-2mm}
\end{figure}

Although many ANN methods have already been proposed,
there are two critical problems of practical
importance that have not received sufficient attention from the research community (\Fref{fig:teaser}).
\begin{itemize} [leftmargin=\ItemizeLeftMargin]
	\item \textit{Subset search} (\Fref{fig:teaser1}):
	Once database vectors are stored, modern ANN systems can run a search efficiently for the \textbf{whole} database.
	Surprisingly, however, almost no systems can run a search over a \textbf{subset} of the database\footnote{
		For example, the state-of-the-art systems Faiss~\cite{web_faiss} and Annoy~\cite{web_annoy} do not provide this functionality. See discussion at \url{https://github.com/facebookresearch/faiss/issues/322},
		\url{https://github.com/spotify/annoy/issues/263}
	}.
	For example, let us consider an image search problem, where the search is formulated as an ANN search over feature vectors.
	We assume that each image also has a corresponding shooting date.
	Given a query image, an ANN system can easily find similar images from the whole dataset.
	However, it is not trivial to find similar images that were taken on a target date (say, May 28 1987).
	Here, the search should not be conducted over the whole dataset, but rather over a subset of the dataset,
	where the subset is specified by identifiers of target images. 
	The straightforward solution is to run the search and check whether or not the results were taken on May 28,
	but this post-checking can be drastically slow, especially if the size of the subset is small.
	Current ANN systems cannot provide a clear solution to this problem.
	
	\item \textit{Performance degradation via data addition} (\Fref{fig:teaser2}):
	So far, the manner in which the search performance
	degrades when items are newly added has not been discussed.
	The number of database items is typically assumed to be provided when an ANN system is built.
	Parameters of the system are usually optimized by taking this number into consideration.
	However, in a practical scenario, new items might be often added to the system.
	Although the performance does not change while the number of new items is small,
	we can ask whether the system remains efficient even after $100\times$ items are newly added.
	To put this another way, suppose that one would like to develop a search system that can handle 1,000,000 vectors in the future,
	but only has 1,000 vectors in the initial stage.
	In such a case, is the search fast even for 1,000 vectors? 
	
\end{itemize}

We develop an ANN system that solves the above two problems, namely \textit{reconfigurable inverted index (Rii)}.
The key idea is extremely simple: storing the data \textbf{linearly}. 
Based on the well-known inverted file with product quantization (PQ) approach (IVFADC)~\cite{tpami_jegou2011},
we design the data layout such that an item can be fetched by its identifier with a cost of $\mathcal{O}(1)$.
This simple but critical modification enables us to search over a subset of the dataset efficiently
by switching to a linear PQ scan if the size of the subset is small.
Owing to this linear layout, the granularity of a coarse assignment step can easily be controlled
by running clustering again over the dataset whenever the user wishes.
This means that the data structure can be adjusted dynamically
after new items are added.

An extensive comparison with state-of-the-art systems, such as Faiss~\cite{web_faiss},
Annoy~\cite{web_annoy}, Falconn~\cite{web_falconn}, and NMSLIB~\cite{web_nmslib},
shows that Rii achieves a comparable performance.
For subset searches and data-addition problems for which the existing approaches do not perform well,
we demonstrate that Rii remains fast in all cases.

Our contributions are summarized as follows.
\begin{itemize}[leftmargin=\ItemizeLeftMargin]
	\item Rii enables efficient searching over a subset of the whole database, regardless of the size of the subset.
	\item Rii remains fast, even after many new items are added, because the data structure
	is dynamically adjusted for the current number of database items.

\end{itemize}

\section{Related Work}
We review existing work
that is closely related to our approach.

\paragraph{Locality-sensitive-hashing}
Locality-sensitive-hashing (LSH)~\cite{scg_datar2004} can be considered as one of the most popular branches of ANN.
Hash functions are designed such that the probability of collision is higher for close points 
than for points that are widely separated.
Using these functions with hash tables,
nearest items can be found efficiently.
Although it has been said that LSH requires a lot of memory and is not accurate compared 
to data-dependent methods,
a recent well-tuned library (FALCONN~\cite{nips_andoni2015, web_falconn}) using multi-probe technology~\cite{vldb_lv2007}
has achieved a reasonable performance.

\paragraph{Projection/tree-based approach}
Space partitioning using a projection or tree
constitutes another significant branch of ANN.
Especially in the computer vision community, 
one of the most widely employed methods is FLANN~\cite{tpami_muja2014}.
Recently, 
the random projection forest-based method
Annoy~\cite{web_annoy} achieved a good performance for million-scale data.

\paragraph{Graph traversal}
Benchmark scores~\cite{sisap_aumuller2017, web_annbenchmark} show that graph traversal-based methods~\cite{is_malkov2014, corr_malkov2016} achieve the current best performance (the fastest with a fixed recall)
when the number of database items is around one million.
These methods first create a graph where each node corresponds to a database item,
which is called a navigable small world.
Given a query, the algorithm starts from a random initial node.
The graph is traversed to the node that is the closest to the query.
In particular, the hierarchical version HNSW~\cite{corr_malkov2016} with
the highly optimized implementation NMSLIB~\cite{sisap_boytsov2013} represents the current state-of-the-art.
The drawback is that it tends to consume memory space, with a long runtime for building the data structure.

\paragraph{Product quantization}
Product quantization (PQ)~\cite{tpami_jegou2011} and
its extensions~\cite{tpami_ge2014, cvpr_norouzi2013, cvpr_babenko2014, eccv_martinez2016, icml_zhang2014, cvpr_zhang2015, cvpr_babenko2015, eccv_douze2016, cvpr_heo2014, eccv_jain2016, iccv_babenko2017, tkde_wang2015}
are popular approaches to handling large-scale data.
Our proposed Rii method also follows this line.
PQ-based methods compress vectors into short memory-efficient codes.
The Euclidean distance between an original vector and compressed code can be efficiently approximated
using a lookup table.
Current billion-scale search systems are usually based on PQ methods,
especially combined with an inverted index-based architecture~\cite{tpami_babenko2015, cvpr_kalantidis2014, tmm_matsui2018, iccv_iwamura2013, cvpr_heo2016, tmm_spyromitros2014, iccv_xia2013}.
Hardware-based acceleration has also recently been
discussed~\cite{vldb_andre2015, icmr_andre2017, kdd_blalock2017, cvpr_wieschollek2016, corr_johnson2017, cvpr_zhang2018, cikm_liu2017}.
An efficient implementation proposed by the original authors is Faiss~\cite{corr_johnson2017, web_faiss}.
An extensive survey is given in~\cite{ite_matsui2018}.


\section{Background: Product Quantization}
In this section, we will review product quantization (PQ)~\cite{tpami_jegou2011}.
PQ compresses vectors into memory efficient short codes. The squared Euclidean distance between
an input vector and the compressed code can be approximated efficiently.
Owing to its memory-efficient form,
PQ played a central role in large-scale ANN systems.

We first describe how to encode a vector. A $D$-dimensional input vector $\mathbf{x} \in \mathbb{R}^D$
is split into $M$ sub-vectors.
Each $D/M$-dimensional sub-vector is compared to $Z$ pre-trained code words, and the identifier
(an integer in $\{1, 2, \dots, Z \}$) of the closest one is recorded.
Using this, $\mathbf{x}$ is encoded as $\mathbf{\bar{x}}$, which is a tuple of $M$ integers:
\begin{equation}
\mathbf{x} \mapsto \mathbf{\bar{x}} = [\bar{x}^1, \dots, \bar{x}^M ]^\top \in \{1, \dots, Z\}^M,
\end{equation}
where the $m$th sub-vector in $\mathbf{x}$ is quantized into $\bar{x}^m$.
We refer to $\mathbf{\bar{x}}$ as a PQ-code for $\mathbf{x}$.
Note that $\mathbf{\bar{x}}$ is represented by $M\log_2 Z$ bits, and we set $Z$ to 256 in order to represent each code using $M$ bytes.

Next, we show how to search over the PQ-codes given
a query vector $\mathbf{q} \in \mathbb{R}^D$.
First, a distance table $\mathbf{A} \in \mathbb{R}^{M \times Z}$ is computed online by comparing the query to the code words.
Here, $A(m, z)$ is the squared Euclidean distance
between the $m$th part of $\mathbf{q}$ and $z$th code word from the $m$th codebook.
The squared Euclidean distance between the query $\mathbf{q}$ and the database vector $\mathbf{x}$ can be
approximately computed using the PQ-code $\mathbf{\bar{x}}$, as follows:
\begin{equation}
d(\mathbf{q}, \mathbf{x})^2 \sim d_A(\mathbf{q}, \mathbf{\bar{x}})^2 = 
\sum_{m=1}^M A(m, \bar{x}^m).
\end{equation}
This is called an asymmetric distance computation (ADC)~\cite{tpami_jegou2011}, and  can be performed efficiently, because only $M$ fetches are required on $\mathbf{A}$. 
A search over $N$ PQ-codes requires $\mathcal{O}(DZ + MN)$.

\section{Reconfigurable Inverted Index}
Now, we introduce our proposed approach: reconfigurable inverted index (Rii).
Let us define a query vector $\mathbf{q}\in\mathbb{R}^D$,
$N$ database vectors $\mathcal{X}=\{ \mathbf{x}_n \in \mathbb{R}^D \}_{n=1}^N$,
and target identifiers $\mathcal{S} \subseteq \{1, \dots, N  \}$.
The subset-search problem is defined to find the $R$ similar items to the query
from the subset of $\mathcal{X}$ specified by $\mathcal{S}$:
\begin{equation}
R\mathchar`-\argmin_{s \in \mathcal{S}} \Vert \mathbf{q} - \mathbf{x}_s \Vert_2^2,
\end{equation}
where the $R\mathchar`-\argmin$ operator finds the $R$ arguments
for which an objective function attains $R$ (sorted) smallest values.
The exact solution can be obtained by a time-consuming direct linear scan.
Our goal is to approximately find nearest items in a fast and memory-efficient manner.
Note that the problem turns out to be a usual ANN search if the whole database is set as the subset:
$\mathcal{S}=\{1, \dots, N\}$.

\subsection{Data Structure}
\label{sec:data_structure}
First, $N$ input database vectors $\mathcal{X}$
are encoded as PQ-codes
$\mathcal{\bar{X}} = \{ \mathbf{\bar{x}}_n \}_{n=1}^N$,
where each $\mathbf{\bar{x}}_n \in \{1, \dots, Z \}^M$.
These PQ-codes are stored \textbf{linearly}, meaning that they are stored in a single long array.
Given an identifier $n$, fetching $\mathbf{\bar{x}}_n$ requires a computational cost of $\mathcal{O}(1)$.

The PQ-codes are clustered into $K$ groups for inverted indexing.
First, $K$ coarse centers $\mathcal{\bar{C}} = \{ \mathbf{\bar{c}}_k \}_{k=1}^K $ are created
by running the clustering algorithm~\cite{mm_matsui2017} on $\mathcal{\bar{X}}$ (or its subset).
Note that each coarse center is also a PQ-code $\mathbf{\bar{c}}_k \in \{1, \dots, Z\}^M$.
Using these coarse centers, the database PQ-codes $\mathcal{\bar{X}}$ are clustered into $K$ groups.
The resulting assignments are stored as posting lists $\mathcal{W}=\{ \mathcal{W}_k \}_{k=1}^K$, where
each $\mathcal{W}_k$ is a set of identifiers of the database vectors whose nearest coarse center is the $k$th one:
\begin{equation}
\mathcal{W}_k = \{ n \in \{1, \dots, N \} | a(n) = k  \}.
\end{equation}
Note that $a(n)$ is an assignment function, that is  defined as $a(n)$ $=$ $\argmin_{k \in \{1, \dots, K\}} d_S(\mathbf{\bar{x}}_n, \mathbf{\bar{c}}_k) $,
where $d_S$ is a symmetric distance function that measures the distance between two PQ-codes~\cite{tpami_jegou2011,mm_matsui2017}.
Finally, we store $\mathcal{\bar{X}}$, $\mathcal{\bar{C}}$, and $\mathcal{W}$ as a data structure for Rii.
The total theoretical memory usage is $(N+K)M\log_2 Z + 32N$ bits if an integer is represented by 32 bits.
We will show in \Sref{sec:exp_comp} that this theoretical value is almost the same as the measured value.

Note that in a typical implementation of the original IVFADC~\cite{tpami_jegou2011} system,
PQ-codes are stored in posting lists (not a single array).
That is, $\{ \mathbf{\bar{x}}_n | a(n) = k\}$ are chunked for each $k$ and then stored.
This would enhance the locality of the data, and improve the cache efficiency when traversing a posting list.
However, the experimental results (\Sref{sec:exp_comp}) showed that this difference is not serious.

\subsection{Search}
We explain how to search for similar vectors using the data structure explained above.
Our system provides two search methods: \textit{PQ-linear-scan} and \textit{inverted-index}.
The former is fast when the size of a target subset is small, and the latter is fast when the size is large.
Depending on the size, the faster method is automatically selected.

A search over a subset of a database is defined as 
a search on target PQ-codes denoted by the target identifiers $\mathcal{S} \subseteq \{1, \dots, N \}$.
Note that we assume that the elements of $\mathcal{S}$ are \textbf{sorted}\footnote{
	A set is denoted by calligraphic font, such as $\mathcal{X}$, 
	and implemented by a single array.
}. 
This is a slightly strong but reasonable assumption.
Because $\mathcal{S}$ is sorted, it can be checked whether or not an item is contained in a set ($n \in \mathcal{S}$) with a cost of $\mathcal{O}(\log_2 |\mathcal{S}|)$ 
using a binary search, where $|\mathcal{S}|$ is the number of elements in $\mathcal{S}$.
Note again that a search over the whole dataset is available by setting $\mathcal{S} = \{1, \dots, N\}$.

\paragraph{PQ-linear-scan}:
Because the database PQ-codes are stored linearly, we can simply pick up target PQ-codes and evaluate the distances to the query.
We call this a PQ-linear-scan.
This is essentially fast if $|\mathcal{S}|$ is small, because only a fraction of vectors are compared.
The pseudocode is presented in \Aref{alg:pq_linear_scan}.

As inputs, the system accepts a query vector $\mathbf{q} \in \mathbb{R}^D$, 
database PQ-codes $\mathcal{\bar{X}}=\{\mathbf{\bar{x}}_n \}_{n=1}^N$,
the number of returned items $R \in \{1, \dots, N \}$,
and the target identifiers $\mathcal{S} \subseteq \{1, \dots, N\}$.
First, a distance table $\mathbf{A}$ is created by comparing a query to code words\footnote{
We intentionally omitted the code words from the pseudocode, for simplicity.} (L1).
This is an online pre-processing step, required for all PQ-based methods.
To store the results, an array of tuples is prepared (L2).
Each tuple consists of (1) an identifier of an item and (2) the distance between
the query and the item.
For each target identifier $s$, the asymmetric distance to the query is computed (L4). 
This distance is then stored in the result array with its identifier $s$,
where the \texttt{PushBack} function is used to append an element to an array (L5).
After all target items have been evaluated, the result array is sorted by the distance (L6).
As we require only the top $R$ results, we use a partial sort algorithm.
Finally, the top $R$ elements are returned,
where the \texttt{Take} function simply picks up the first several elements (L7).
Note that $\mathcal{W}$ and $\mathcal{\bar{C}}$ are not required for the search.

Let us analyze the computational cost. The creation of a distance table requires $\mathcal{O}(DZ)$,
and a comparison to $|\mathcal{S}|$ items requires $\mathcal{O}(M|\mathcal{S}|)$.
Partial sorting requires $\mathcal{O}(|\mathcal{S}|\log_2 R)$ on average\footnote{
	This cost 
	comes from the heap sort-based implementation used in the \texttt{partial\_sort} function in C++ STL.
	Another option is to pick up the $k$ smallest items and only sort these.
	This leads to  $\mathcal{O}(|\mathcal{S}| + R \log_2 R)$.
	We used the former in this paper
	because we empirically determined that the former is faster in practice, especially
	when $R$ is small.
}.
Their sum leads to a final average cost (\Tref{tbl:comp_cost}).
It is clear that the computation is efficient if $|\mathcal{S}|$ is small.
As the cost depends on $|\mathcal{S}|$ linearly, a PQ-linear-scan becomes inefficient if $|\mathcal{S}|$ is large.
Note that if the search target is the whole dataset, $|\mathcal{S}|$ is replaced by $N$.

\begin{algorithm}[tb]
	\KwIn{$\mathbf{q}\in\mathbb{R}^D$, \ \ \ \ \ \ \ \ \ \ \ \ \ \codecomment{\# Query} \newline
		$\mathcal{\bar{X}}=\{\mathbf{\bar{x}}_n \}_{n=1}^N$, \ \ \ \ \codecomment{\# Database PQ-codes} \newline
		$R \in \{1, \dots, N \}$, \ \codecomment{\# \# of returned items} \newline
		$\mathcal{S} \subseteq \{1, \dots, N\}$ \ \codecomment{\# Target subset identifiers}}
	\KwOut{$\mathcal{U} = \{ \mathbf{u}_r \}_{r=1}^R$ s.t. $\mathbf{u}_r = [n_r, d_r] \in \{1, \dots, N\} \times \mathbb{R}$ \newline
	\codecomment{\# $n_r$: r-th identifier. $d_r$: r-th distance}.}
	$\mathbf{A} \gets \mathtt{CompareCodewords}(\textbf{q})$ \ \codecomment{\# Distance table} \\
	$\mathcal{U} \gets \emptyset$ \ \codecomment{\# Array of tuples (id, distance)} \\
	\For{$s \in \mathcal{S}$}{
		$d \gets \sum_{m=1}^M A(m, \bar{x}_s^m)$ \\
		$\mathtt{PushBack}(\mathcal{U}, [s, d])$
	}
	$\mathtt{PartialSort}(\mathcal{U}, R)$ \ \codecomment{\# Sort by distance} \\
	\Return{$\mathtt{Take}(\mathcal{U}, R)$} \ \codecomment{\# Top R}
	\caption{\texttt{PQLinearScan}}
	\label{alg:pq_linear_scan}
\end{algorithm}

\begin{table}
	\begin{center}
		\caption{
			The average computational complexity for each operation.
			The range for each variable used in this paper:
			$96 \le D \le 960$, $Z=256$, $8 \le M \le 240$, $10^6 \le N \le 10^9$,
			$1 \le R \le 100$, $10^2 \le |\mathcal{S}| \le 5 \times 10^5$,
			$10^3 \le K \le 3.2\times10^4$, $10^3 \le L \le 3.2 \times 10^4$.
		}
		\vspace{-3mm}
		\label{tbl:comp_cost}				
		\scalebox{0.8}{			
			\begin{tabular}{@{}ll@{}} \toprule %
				Operation & Computational complexity \\ \midrule
				\texttt{PQLinearScan} & \\
				\ \ - whole data  & $\mathcal{O}(DZ + MN + N\log_2 R)$ \\ 
				\ \ - susbet $(\mathcal{S})$ & $\mathcal{O}(DZ + M |\mathcal{S}| + |\mathcal{S}|\log_2 R)$ \\
				\texttt{InvertedIndex} & \\
				\ \ - whole data  & $\mathcal{O} \left (DZ + KM + K\log_2 \frac{KL}{N} + LM + L\log_2 R \right )$ \\ 
				\ \ - susbet $(\mathcal{S})$ & $ \mathcal{O} \left (DZ + KM + K\log_2 \left ( \min \left ( \frac{KL}{|\mathcal{S}|}, K \right ) \right )
				+ \frac{LN}{|\mathcal{S}|} \log_2 |\mathcal{S}| + LM + L\log_2 R \right ) $ \\
				\bottomrule
			\end{tabular}
		}
	\end{center}
	\vspace{-2mm}		
\end{table}

\paragraph{Inverted-index}:
The other search method is inverted-index. 
Because the database items are preliminarily clustered as explained in \Sref{sec:data_structure},
we can simply evaluate items that are in the same/close clusters to the query.
This drastically boosts the performance
if the number of the target identifiers is large.

We show the pseudo-code in \Aref{alg:ivf}.
Inverted-index takes three additional inputs: posting lists $\mathcal{W}$, coarse centers $\mathcal{\bar{C}}$,
and the number of candidates $L$.
Note that $L$ candidates will be selected and evaluated in the final step.
This means that $L$ is a runtime parameter that controls the trade-off between the accuracy and runtime.

To search, a distance table is first created in the same manner as for PQ-linear-scan (L1).
The search steps consists of two blocks.
First, the closest clusters to the query are found (L2-6).
Then, the items inside the clusters are evaluated (L7-16).

To find the closest clusters, an array of tuples is created (L2).
For each coarse center ($\mathbf{\bar{c}}_k)$, the distance from the query is computed (L4).
The results are stored in the array (L5).

Next, we run partial sort on the array to find the closest clusters to the query (L6).
Here, the target number of the partial sort (the number of postings lists to be focused)
is set as $\left \lceil \frac{KL}{|\mathcal{S}|} \right \rceil$, which is determined as follows.
Because the target identifiers are of size $|\mathcal{S}|$, where the total number of identifiers is $N$,
the probability of any item being a target identifier is $|\mathcal{S}|/N$ on average.
Because our purpose here is to select $L$ target items as candidates of the search,
the required number of items to traverse is $L / (|\mathcal{S}|/N) = LN/|\mathcal{S}|$.
To traverse $LN/|\mathcal{S}|$ items, we need to focus on $(LN/|\mathcal{S}|) / (N/K) = KL/|\mathcal{S}|$ posting lists,
because the average number of items per posting list is $N/K$.
This implies that we need to select the nearest $\left \lceil \frac{KL}{|\mathcal{S}|} \right \rceil$ posting lists.
Note that if $K < \frac{KL}{|\mathcal{S}|}$, we simply replace the value by $K$, because this performs a full sort of the array ($\mathcal{O}(K\log_2 K)$).

The selected posting lists are then evaluated. 
A score array is prepared (L7).
For each closest posting list (L8), identifiers in the posting list are traversed (L9).
If an identifier is not included in the target identifier $\mathcal{S}$, then
this item is simply ignored (L10-11).
Note that if the search is for the whole dataset ($\mathcal{S}=\{1, \dots, N\}$),
any item $n$ is always included in $\mathcal{S}$,
thus we remove L10-11.

For a selected identifier $n$, the identifier and the distance to the query are recorded 
in the same manner as for the PQ-linear-scan (L12-13).
If the size of the score array ($|\mathcal{U}|$) reaches the parameter $L$, then
the top $R$ results are selected and returned (L14-16).

The computational cost is summarized as follows. 
After the code creation with $\mathcal{O}(DZ)$, the comparison to $K$ coarse centers requires $\mathcal{O}(KM)$.
Partial sort requires $\mathcal{O}(K \log_2 (KL/|\mathcal{S}|))$.
The number of items to be traversed is 
$\mathcal{O}(LN/|\mathcal{S}|)$. We can check whether or not each item is included in $\mathcal{S}$ 
using a binary search, requiring $\mathcal{O}(\log_2 |\mathcal{S}|)$.
This leads to $\mathcal{O}(LN/|\mathcal{S}| \cdot \log_2 |\mathcal{S}|)$ in total.
The number of items that are actually evaluated is $L$, and so $\mathcal{O}(LM)$ of the cost is required.
Finally, the top $R$ are selected using the partial sort, requiring $\mathcal{O}(L\log_2 R)$.
\Tref{tbl:comp_cost} summarizes the computational cost.
Inverted-index is fast when $|\mathcal{S}|$ is sufficiently large, but is slow if $|\mathcal{S}|$ is small.
This is highlighted in the term $LN/|\mathcal{S}| \log_2 |\mathcal{S}|$, where this term becomes dominant if 
$|\mathcal{S}|$ is small.

Note that although there appear to be several input parameters for inverted-index,
all of them except $L$ are usually decided deterministically.
$L$ is the only parameter the user needs to decide.
Our initial setting is the average length of a posting list, $L=N/K$.
This means that the system traverses one posting list on average.
This is a fast setting, and users can change this if they require more accuracy,
as $L=2N/K, 3N/K, ...$.

\begin{algorithm}[tb]
	\KwIn{$\mathbf{q}\in\mathbb{R}^D$, \ \ \ \ \ \ \ \ \ \ \ \ \ \codecomment{\# Query} \newline
		$\mathcal{\bar{X}}=\{\mathbf{\bar{x}}_n \}_{n=1}^N$, \ \ \ \ \codecomment{\# Database PQ-codes} \newline
		$\mathcal{W}=\{ \mathcal{W}_k \}_{k=1}^K$, \codecomment{\# Posting lists} \newline
		$\mathcal{\bar{C}}=\{ \mathbf{\bar{c}}_k \}_{k=1}^K$, \ \ \ \ \ \codecomment{\# Coarse centers} \newline
		$R \in \{1, \dots, N \}$, \ \ \codecomment{\# \# of returned items} \newline
		$\mathcal{S} \subseteq \{1, \dots, N\}$, \ \codecomment{\# Target subset identifiers} \newline
		$L \in \{1, \dots, N \}$ \ \ \ \codecomment{\# \# of candidates}}
	\KwOut{$\mathcal{U} = \{ \mathbf{u}_r \}_{r=1}^R$ s.t. $\mathbf{u}_r = [n_r, d_r] \in \{1, \dots, N\} \times \mathbb{R}$ \newline
		\codecomment{\# $n_r$: r-th identifier. $d_r$: r-th distance}.}
	$\mathbf{A} \gets \mathtt{CompareCodewords}(\textbf{q})$ \ \codecomment{\# Distance table} \\
	$\mathcal{T} \gets \emptyset$ \ \codecomment{\# Array of tuples (id, distance)} \\
	\For{$k \in \{1, \dots, K\}$}{
		$d_0 \gets \sum_{m=1}^M A(m, \bar{c}_k^m)$ \\	
		$\mathtt{PushBack}(\mathcal{T}, [k, d_0])$
	}
	$\mathtt{PartialSort} \left (\mathcal{T}, \left \lceil \frac{KL}{|\mathcal{S}|} \right \rceil \right )$ \ \codecomment{\# Sort by distance} \\
	$\mathcal{U} \gets \emptyset$ \ \codecomment{\# Array of tuples (id, ditance)} \\
	\For{$[k, d_0] \in \mathcal{T}$}{
		\For{$n \in \mathcal{W}_k$}{
			\If{$n \notin \mathcal{S}$}{
				\textbf{continue}
			}
			$d \gets \sum_{m=1}^M A(m, \bar{x}_n^m)$ \\
			$\mathtt{PushBack}(\mathcal{U}, [n, d])$ \\
			\If{$|\mathcal{U}| = L$}{
				$\mathtt{PartialSort}(\mathcal{U}, R)$ \ \codecomment{\# Sort by distance} \\
				\Return{$\mathtt{Take}(\mathcal{U}, R)$} \ \codecomment{\# Top R}
			}
		}
	}
	\caption{\texttt{InvertedIndex}}
	\label{alg:ivf}
\end{algorithm}

\begin{figure}
	\includegraphics[width=1.0\linewidth]{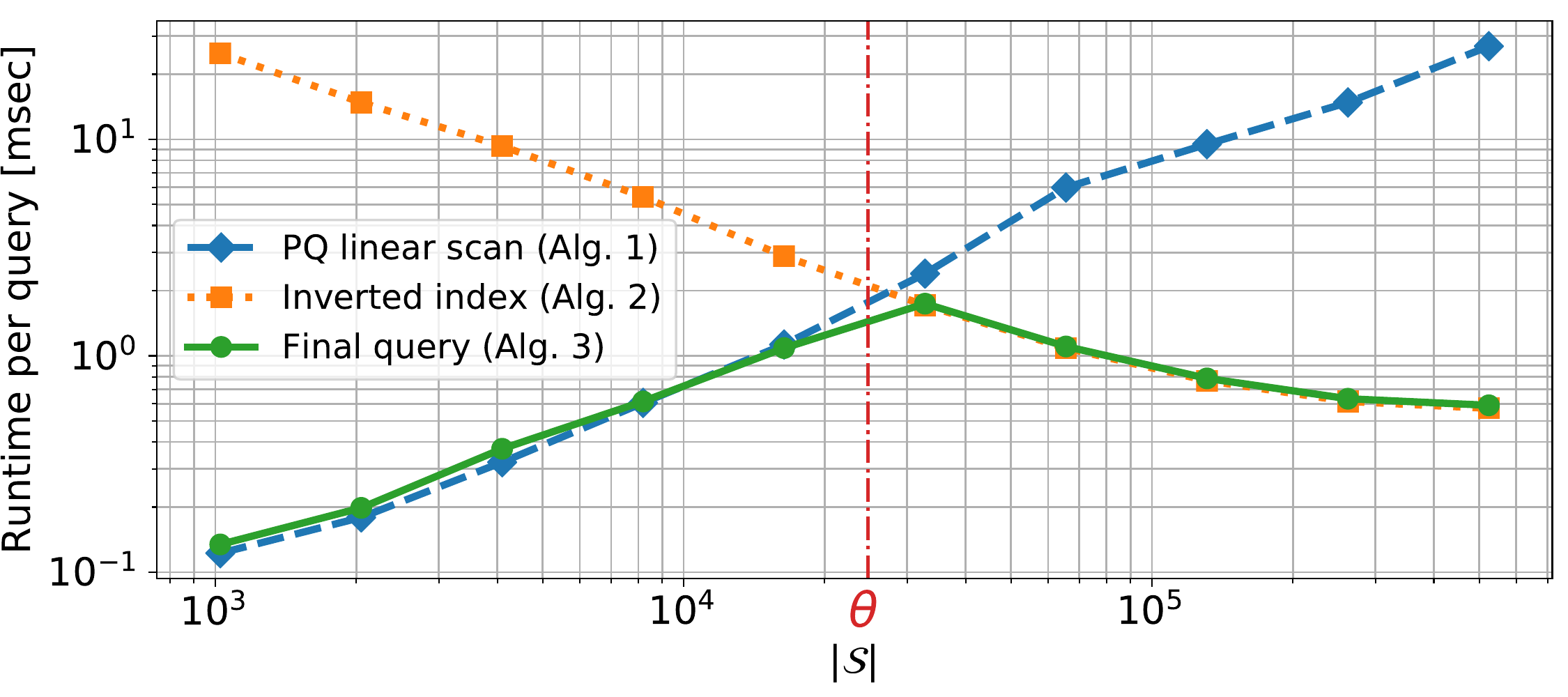}
	\vspace{-7mm}
	\caption{
		Comparison of PQ-linear-scan, inverted-index, and the final query algorithm.
		Runtime per query for the SIFT1M dataset with various sizes of target identifiers is plotted.
		Note that $L=K=1000, R=1, \theta=24743$.
	}
	\label{fig:linear_ivf}
	\vspace{-5mm}
\end{figure}

\paragraph{Selection}:
The final query algorithm is described in \Aref{alg:query}.
Given inputs, the system automatically determines the query method as either PQ-linear-scan or inverted-index.
This decision is based on the threshold value $\theta$ for the number of target identifiers (L1).
Owing to this flexible switching, we can always achieve a fast search with a single Rii data structure
($\mathcal{\bar{X}}$, $\mathcal{W}$, and $\mathcal{\bar{C}}$),
regardless of the sizes of the target identifiers ($|\mathcal{S}|$).
\Fref{fig:linear_ivf} highlights the relations among the three query algorithms.

Note that it is not trivial to set the threshold $\theta$ deterministically,
because it depends on several parameters, such as $M$ and $L$.
To find the best threshold, 
we simply run the search with several parameter combinations
when the data structure is constructed.
Based on the result, we fit a 1D line in the parameter space,
and finally obtain the best threshold.
See the supplementary material for more details.
This works almost perfectly, as shown in \Fref{fig:linear_ivf}.
This thresholding does not require any additional runtime cost for the search phase.

\begin{algorithm}[tb]
	\KwIn{$\mathbf{q}$,
		$\mathcal{\bar{X}}$,
		$\mathcal{W}$,
		$\mathcal{\bar{C}}$,
		$R$,
		$\mathcal{S}$,
		$L$ \newline
		\codecomment{\# See the definitions in Alg.~\ref{alg:ivf}}}
	\KwOut{$\mathcal{U} = \{ \mathbf{u}_r \}_{r=1}^R$ s.t. $\mathbf{u}_r = [n_r, d_r] \in \{1, \dots, N\} \times \mathbb{R}$}
	\uIf{$|\mathcal{S}| < \theta$ }{
		\Return{$\mathtt{PQLinearScan}(\mathbf{q}, \mathcal{\bar{X}}, R, \mathcal{S})$} \ \codecomment{\# Alg.~\ref{alg:pq_linear_scan}}
	}\Else{
		\Return{$\mathtt{InvertedIndex}(\mathbf{q}, \mathcal{\bar{X}}, \mathcal{W}, \mathcal{\bar{C}}, R, \mathcal{S}, L)$} \ \codecomment{\# Alg.~\ref{alg:ivf}}
	}
	\caption{\texttt{Query}}
	\label{alg:query}
\end{algorithm}

\subsection{Reconfiguration}
\label{sec:reconfigure}
Here, we introduce a reconfigure function that enables us to search efficiently
even if a large number of vectors are newly added.
As discussed in \Sref{sec:intro}, typical ANN systems are first optimized
to achieve fast searching for $N$ items. If new items are added later, such systems might become slow.
For example, IVFADC requires an initial decision on the number of space partitions $K$.
The selection of $K$ is sensitive and critical to the performance.
A standard convention\footnote{\url{https://github.com/facebookresearch/faiss/wiki/Index-IO,-index-factory,-cloning-and-hyper-parameter-tuning}}
is to set $K=\sqrt{N}$.
On the other hand, $K$ cannot be changed later. The system could become slower
if $N$ changes significantly.
In other words, we must decide $K$ even if the final database size $N$ is not known,
which sometimes frustrates users.

Unlike these existing methods, Rii provides a reconfigure function.
If the search becomes slow because of newly added items,
coarse centers and assignments are updated by simply running clustering again.
The system is automatically optimized to achieve the fastest search for the current number of 
database items.

\paragraph{Data addition}
Let us first explain how to add a new item. Given a new PQ-code $\mathbf{\bar{y}}$,
the database PQ-codes $\mathcal{\bar{X}}=\{\mathbf{\bar{x}}_n\}_{n=1}^N$
are updated using \texttt{PushBack}($\mathcal{\bar{X}}$, $\mathbf{\bar{y}}$).
A corresponding posting list is also updated by \texttt{PushBack}($\mathcal{W}_{a(N+1)}, N+1$).
Then, searching can be performed without any modifications, but it may be slower if many items are added.
This is because the length of each posting list ($|\mathcal{W}_k|$) can become too long, making the traversal inefficient.

\paragraph{Reconfigure}
If the search becomes slow, a reconfigure function can be called (\Aref{alg:reconfigure}).
This function takes the database PQ-codes $\mathcal{\bar{X}}$ and a new number of coarse space partitions $K'$ as inputs.
Again, $K'$ is typically set as $\sqrt N$ for the new $N$.
The outputs are updated posting lists and coarse centers.
First, the updated coarse centers are computed by running clustering over the PQ-codes using PQk-means~\cite{mm_matsui2017} (L1).
PQk-means efficiently puts the input PQ-codes into several clusters, without decoding the codes for the original $D$-dimensional vectors.
Note that clustering can be run for a subset of $\mathcal{\bar{X}}$ to make this fast.
We set the upper limit of the codes to be clustered as $\min(N, 100K')$.
After new coarse centers are obtained, the posting lists are created by simply
finding the nearest center for each PQ-code (L2-4).

The advantage of the reconfigure function is that it can be called whenever the user wishes. 
The results are deterministic for $K'$, because this just runs the clustering over the codes.
We will show in \Sref{sec:exp_reconfigure} that this reconfigure function is especially useful when the database size drastically changes.
Another way of looking at this is that  
we do not need to know the final number of database items when the index structure is built.
This is a clear advantage over IVFADC-based methods.
In a practical scenario, it will often occur that the number of database items cannot be decided when the system is created.
Even in such cases, IVFADC must decide the parameters. This would lead to a suboptimal performance.

\begin{algorithm}[tb]
	\KwIn{$\mathcal{\bar{X}}=\{\mathbf{\bar{x}}_n \}_{n=1}^N$, \ \ \ \ \ \ \ \ \codecomment{\# Database PQ-codes} \newline
		$K'\in\{1, \dots, N\}$ \ \ \ \ \codecomment{\# \# of coarse centers}}
	\KwOut{$\mathcal{W}=\{\mathcal{W}_k \}_{k=1}^{K'}$, \ \codecomment{\# Updated posting list} \newline
		$\mathcal{\bar{C}}=\{ \mathbf{\bar{c}}_k \}_{k=1}^{K'}$ \ \ \ \ \ \ \ \codecomment{\# Updated coarse centers} }
	$\mathcal{\bar{C}} \gets \mathtt{PQkmeans}(\mathcal{\bar{X}}, K')$ \codecomment{\# Clustering on PQ-codes \cite{mm_matsui2017}} \\
	$\mathcal{W} \gets \emptyset$ \\
	\For{$k \in \{1, \dots, K'\}$}{
		$\mathcal{W}_k \gets \{ n \in \{1, \dots, N \} | a(n) = k  \}$
	}
	\Return{$\mathcal{W}, \mathcal{\bar{C}}$}
	
	\caption{\texttt{Reconfigure}}
	\label{alg:reconfigure}
\end{algorithm}

\subsection{Connection to IVFADC}
The data structure proposed above is similar to the original IVFADC~\cite{tpami_jegou2011}, but has the following fundamental differences.
\begin{itemize}[leftmargin=\ItemizeLeftMargin]
	\item In Rii, each vector is encoded directly, whereas IVFADC encodes a residual between an input vector and a coarse center.
	This makes the accuracy of Rii slightly inferior to that of IVFADC (see \Sref{sec:exp_comp}), but enables us to store
	PQ-codes linearly.
	\item In Rii, PQ-codes are stored linearly, and their identifiers are stored in posting lists.
	In IVFADC, both PQ-codes and identifiers are stored in posting lists.
	This simple modification enables us to run the PQ-linear scan without any additional operations.
	\item In IVFADC, coarse centers are a set of $D$-dimensional vectors, whereas coarse centers in Rii are PQ-codes.
	The advantage of this is that the reconfigure steps become considerably fast with PQk-means.
	The limitation is that this might decrease the accuracy, but the experimental results show that this degradation is not serious (\Sref{sec:exp_comp}).
\end{itemize}

\subsection{Advanced Encoding}
There exist advanced encoding methods for PQ, such as optimized product quantization (OPQ)~\cite{tpami_ge2014, cvpr_norouzi2013},
additive quantization (AQ)~\cite{cvpr_babenko2014, eccv_martinez2016}, and
composite quantization (CQ)~\cite{icml_zhang2014, cvpr_zhang2015}.
Although state-of-the-art accuracy has been achieved by AQ or CQ,
it is widely known that they are more complex and time consuming.
Therefore, we did not incorporate AQ and CQ in our system.

On the other hand, OPQ provides a reasonable trade-off (slightly slow but with a high accuracy).
In OPQ, a rotation matrix is preliminarily
trained to minimize the error. 
In the search phase, an input vector is first rotated with the matrix. 
The remaining process is exactly the same as PQ.
We will show the results of OPQ in \Sref{sec:exp_comp}.

\section{Evaluations}
All experiments were performed on a server with a 3.6 GHz
Intel Xeon CPU (six cores, 12 threads) and 128 GB of RAM.
For a fair comparison, we employed a single-thread implementation for the search.
Rii is implemented by C++ with a Python interface,
All source codes are publicly available\footnote{\url{https://github.com/matsui528/rii}}

\subsection{Datasets}
The various methods were evaluated using the following datasets:
\begin{itemize} [leftmargin=\ItemizeLeftMargin]
	\item SIFT1M~\cite{icassp_jegou2011} consists of 128D SIFT feature vectors extracted from several images.
	It provides 1,000,000 base, 10,000 query, and 100,000 training vectors.
	\item GIST1M~\cite{icassp_jegou2011} consists of 960D GIST feature vectors extracted from several images.
	It provides 1,000,000 base, 1,000 query, and 500,000 training vectors.
	\item Deep1B~\cite{cvpr_babenko2016} consists of 96D deep features
	extracted from the last FC layer of GoogLeNet~\cite{cvpr_szegedy2015} for one billion images.
	It provides 1,000,000,000 base, 10,000 query, and 1,000,000 (we used the top 1M from the whole training branch) training vectors.
\end{itemize}

The code words of Rii and Faiss were preliminarily trained using the training data.
The search is conducted over the base vectors.

\subsection{Methods}
We compare our Rii method with the following existing methods:
\begin{itemize} [leftmargin=\ItemizeLeftMargin]
	\item Annoy~\cite{web_annoy}: A random projection forest-based system.
	Because Annoy is easy to use (fewer parameters, intuitive interface, no training steps, and easy IO with a direct mmap design),
	it is the baseline for million-scale data.
	\item FALCONN~\cite{web_falconn}: Highly optimized LSH~\cite{nips_andoni2015}.
	FALCONN is regarded as a representative state-of-the-art LSH-based method.
	\item NMSLIB~\cite{web_nmslib}: Highly optimized ANN library with the support of non-metric spaces~\cite{sisap_boytsov2013}.
	This library includes several algorithms, and we used Hierarchical Navigable Small World (HNSW)~\cite{is_malkov2014, corr_malkov2016}
	in this study. NMSLIB with HNSW is the current state-of-the-art for million-scale data~\cite{sisap_aumuller2017, web_annbenchmark}.
	\item Faiss~\cite{web_faiss}: A collection of highly-optimized PQ-based methods. This library includes IVFADC~\cite{tpami_jegou2011}, OPQ~\cite{tpami_ge2014}, inverted multi-index~\cite{tpami_babenko2015},
	and polysemous codes~\cite{eccv_douze2016}. Some of these are implemented using the GPU as well~\cite{corr_johnson2017}.
	In particular, we compared Rii with the basic IVFADC, which is one of the fastest options.
	Note that only Faiss and Rii can handle billion-scale data, because PQ-based methods are memory efficient.
\end{itemize}

\subsection{Subset Search}
\label{sec:exp_subset}
We first present the results for searching over a subset of the whole database.
This is the main function that the proposed Rii method provides.
The conclusion is that Rii always remains fast,
whereas existing methods become considerably slower, especially if 
the size of the target subset is small.
We first explain the task, and then introduce a post-checking module through which existing methods can conduct a subset search. Finally,
we present the results.

\paragraph{Task}
The task is defined as follows.
We randomly select integers from $\{1, \dots, N\}$, sort them, and construct the target indices $\mathcal{S}\subseteq \{1, \dots, N\}$.
For each query, we run the search and find the top-$R$ results.
All the results must be members of $\mathcal{S}$.
The runtime per query was reported with several combinations of $\mathcal{S}$ and $R$.
The evaluation was conducted using the SIFT1M dataset ($N=10^6$), with $R \in \{1, 10, 100\}$.

\paragraph{Post-checking module}
Because none of the existing methods provide a subset search functionality, 
we implemented a straightforward post-checking module
in order to enable the existing methods to perform a subset search.
\Aref{alg:post_check} shows the pseudocode.
This module takes a query function $Q$, a query vector $\mathbf{q}$, target identifiers $\mathcal{S}$, and the number of returned items $R$ as inputs.
The query function $Q$ returns the identifiers of the $R$ closest items, given $\mathbf{q}$ and $R$.
This $Q$ is an existing method such as Annoy.
First, the output identifier set is prepared (L1).
The number of returned items for each iteration, $r$, is first initialized (L2).
Then, the search begins with an infinite loop.
The top-$r$ items are searched using $Q$, and the results are stored in the temporal buffer $\mathcal{T}$ (L4).
For each identifier $n$ in $\mathcal{T}$, 
if $n$ has already been checked, the loop continues (L6-7).
This is actually achieved by starting a for loop with some offsets over $\mathcal{T}$, so that the first already-checked elements up to a certain number
are not traversed.
If $n$ is included in $\mathcal{S}$, we store it in the output set $\mathcal{U}$ (L8-9).
The algorithm finishes if the enough ($R$) items are found (L10-11).
If an insufficient number of items are found, then $r$ is updated to a larger number by simply multiplying
a constant value (L12).
The search continues with the updated $r$ until $R$ items are found.

With this module, searching over a target subset is made available for the existing methods.
Note that $Q$ cannot always return $r$ items when $r$ is large.
This depends on the design of the query function, and some
methods have a limit on $r$ in order not to make the search too slow.
We found that FALCONN and NMSLIB do not return $r$ items if $r$ is large.
Therefore, we compared Rii with Annoy using the post-checking module (Annoy + PC).

\begin{algorithm}[tb]
	\KwIn{
		$Q$, \ \ \ \ \ \ \ \ \ \ \ \ \ \ \ \ \ \ \ \ \ \ \ \ \codecomment{\# Query function} \newline 
		$\mathbf{q} \in \mathbb{R}^D$, \ \ \ \ \ \ \ \ \ \ \ \ \ \ \ \ \codecomment{\# Query vector} \newline
		$\mathcal{S} \subseteq \{1, \dots, N\}$, \ \ \  \codecomment{\# Target subset identifiers} \newline
		$R \in \{1, \dots, N \}$, \ \ \ \ \codecomment{\# \# of returned items}
	}
	\KwOut{$\mathcal{U} \subseteq \mathcal{S}$ \ \codecomment{\# $\mathcal{U}$ is sorted}}
	$\mathcal{U} \gets \emptyset$ \ \codecomment{\# An array of integers} \\
	$r \gets R$ \\
	\While{1}{
		$\mathcal{T} \gets \mathrm{Q}(\mathbf{q}, r)$  \ \codecomment{\# Return top-$r$ results} \\
		\For{$n \in \mathcal{T}$}{
			\If{$n$ has been already checked}{
				\textbf{continue} 
			}
			\If{$n \in \mathcal{S}$}{
				$\mathtt{PushBack}(\mathcal{U}, n)$
			}
			\If{$|\mathcal{U}| = R$}{
				\Return{$\mathcal{U}$}
			}
		}
	$r \gets r\times 5$ \ \codecomment{\# User defined constant value}
	}		
	\caption{Post-checking module for existing methods.}
	\label{alg:post_check}
\end{algorithm}

\begin{figure}
	\includegraphics[width=1.0\linewidth]{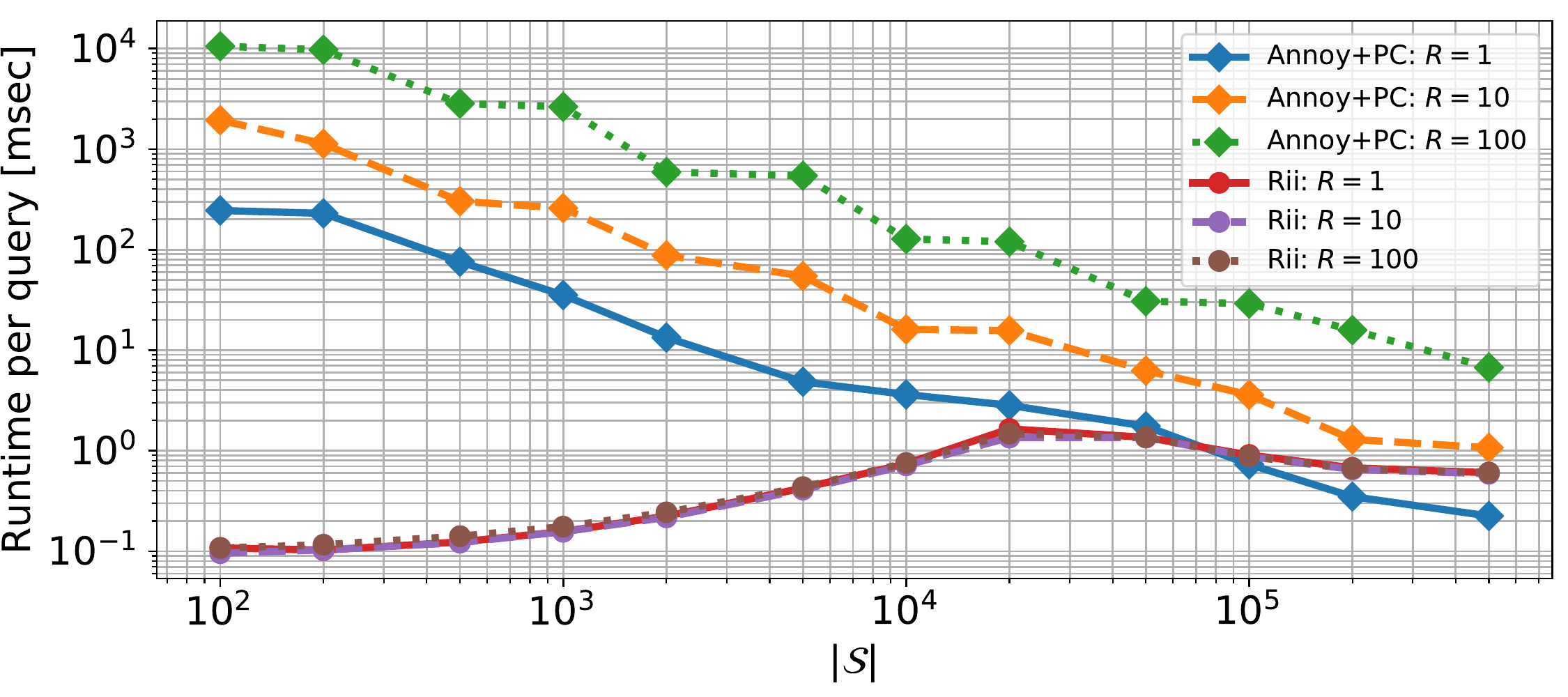}
	\caption{
		Subset search using the SIFT1M dataset over 10 queries.
		Note that $K = L = 1000, M=64$.
	}
	\label{fig:subset}
	\vspace{-5mm}
\end{figure}

\paragraph{Results}
\Fref{fig:subset} illustrates the results.  We point out the following:
\begin{itemize} [leftmargin=\ItemizeLeftMargin]
	\item Rii was fast under all conditions (less than 2 ms/query).
	We can conclude that Rii was stable and effective for the subset-search.
	\item As with IVFADC, Rii is robust against $R$.
	\item Annoy + PC became drastically slow for small $|\mathcal{S}|$,
	which is further highlighted when $R$ is large. 
	This is an obvious result, because the while loop (L3 in \Aref{alg:post_check}) must be repeated several times
	for large $r$.
	Here, $r$ can be even $N$.
	ANN systems are usually not designed to handle such $r$ values.	
\end{itemize}

\subsection{Robustness Against Data Addition}
\label{sec:exp_reconfigure}
We describe the experiments for our other main function, reconfigure.
The conclusion is that Rii becomes fast by using reconfigure,
even after many new vectors are added.
First, the task is explained, then the results are presented.
Here, we used the Deep1B dataset to demonstrate the robustness against 
billion-scale data.

\paragraph{Task}
The index is first constructed using $N=10^6$ vectors with $K=\sqrt{N}=10^3$,
and then the runtime is evaluated.
Next, new items are added to the index, so that the final $N$ becomes $10^7$.
Then, the runtime is evaluated in two ways: (1) a search is performed with $K=10^3$, and
(2) the data structure is updated using the reconfigure function with $K=\sqrt{10^7}$, and
then the search is conducted.
We run this experiment with the final $N$ as $10^7$, $10^8$, and $10^9$.

\begin{figure}
	\includegraphics[width=1.0\linewidth]{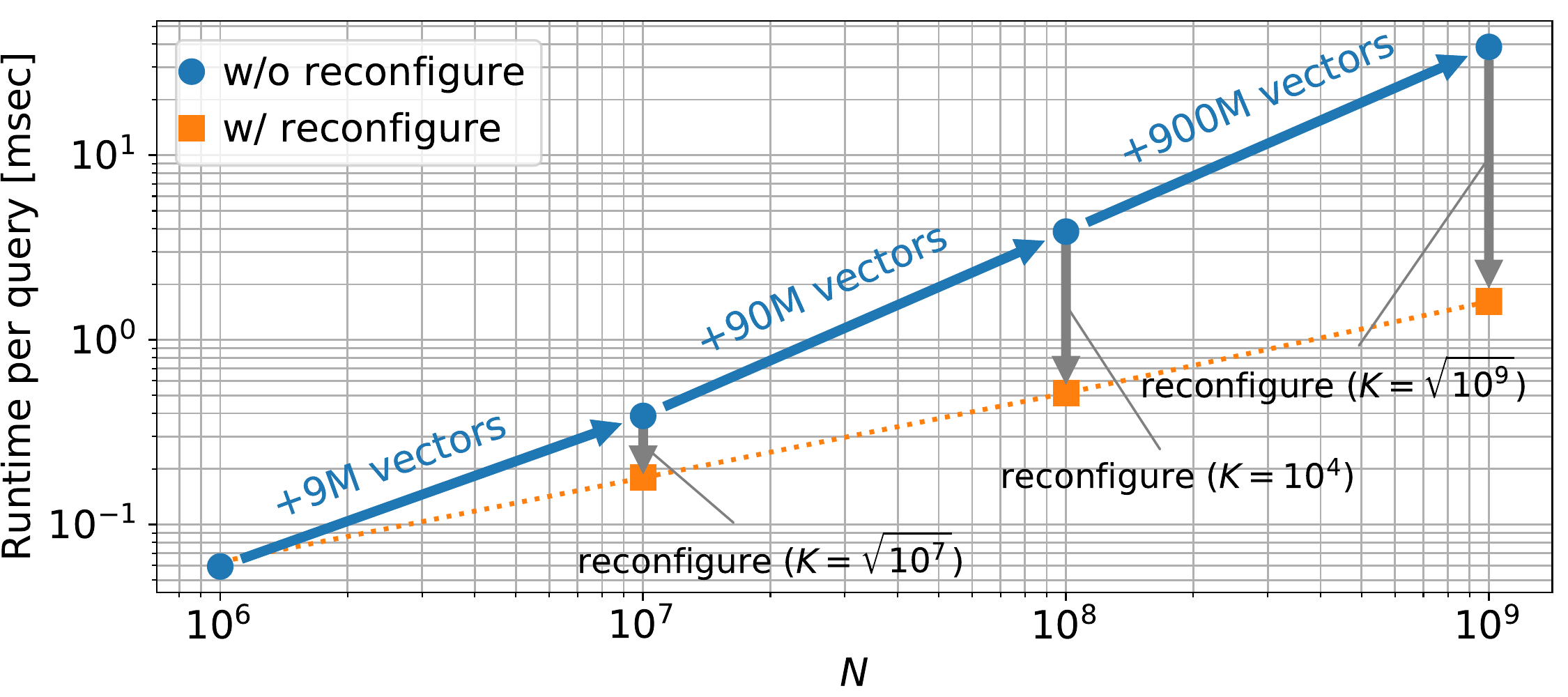}
	\caption{
		The runtime performance with and without the reconfigure function
		over the Deep1B dataset, where $R=1$, $M=8$, and $L=N/K$.
	}
	\label{fig:reconfigure}
	\vspace{-5mm}
\end{figure}

\paragraph{Results}
\Fref{fig:reconfigure} illustrates the result.
It is clear that the search becomes dramatically faster after the reconfigure function is called.
For example, if the user keeps the same data structure after $99M$ new items are added,
the search takes an average of 3.9 ms. This can be made $7.8\times$ faster after applying the reconfigure function.

Most importantly, because the data structure can be always adjusted for the new $N$,
the user need not face the burden of selecting $K$ when the system is constructed. 
This is a clear advantage over the other existing methods.
Note that the runtime for adding $9\times10^6$ vectors was 109 s, and that of the reconfigure function
with $K=\sqrt{10^7}$ was 111 s. These times can be considered moderate.

\subsection{Comparison with Existing Methods}
\label{sec:exp_comp}

Finally, we compare Rii (and its variant Rii-OPQ) with Annoy, FALCONN, NMSLIB (HNSW), and Faiss (IVFADC),
using SIFT1M and GIST1M.
The conclusion is that our Rii method achieved a comparable performance to the state-of-the-art system Faiss.
Note that the searches were conducted over the whole datasets.

The accuracy was measured using Recall@1,
which measures the fraction of queries for which the ground truth nearest neighbor
is returned within the top-1 result.
The average Recall@1 over the query set is reported.
We evaluated the methods with several parameter combinations,
and report the results with a fixed Recall@1 (0.65 for SIFT1M and 0.5 for GIST1M)
for a fair comparison.
Because the ranges of some parameters are discrete, we cannot achieve an exact target Recall@1.
Thus, the target Recall@1 was selected as best as possible as a value that all methods can achieve.

The disk consumption of the index data structure is also reported.
This was measured by storing the data structure on the disk and checking its size in bytes.
Note that the runtime (peak-time) memory consumption is the more important measure,
but measuring the peak-time memory usage
is not always stable, and can vary depending on the computer.
Thus, we report the disk space instead, which is reproducible and strongly related to the memory consumption.
The runtime of building the data structure is also reported.

\begin{table*}
	\begin{center}
		\caption{
			Comparison to existing methods using SIFT1M/GIST1M. Note that $R=1$ for all methods.
			Unless explicitly denoted, we adopt the default parameters for each method. The bold fonts indicate the best scores among the methods.
		}
		\label{tbl:million_compare}				
		\begin{tabular}{@{}lllllll@{}} \toprule %
			Dataset & Method & Parameters & Recall@1 (fixed) & Runtime/query & Disk space & Build time \\ \midrule
			\multirow{6}{*}{SIFT1M}	& Annoy~\cite{web_annoy} & $n_{\mathrm{trees}}=2000$, $k_{\mathrm{search}}=400$ & 0.67 & 0.18 ms & 1703 MB & 899 sec \\ 
			& FALCONN~\cite{web_falconn, nips_andoni2015} & $n_{\mathrm{probes}}=16$ & 0.63 & 0.87 ms & - & \textbf{1.8 sec} \\ 
			& NMSLIB (HNSW)~\cite{web_nmslib, sisap_boytsov2013, corr_malkov2016} & $\mathrm{efS}=4$ & 0.67 & \textbf{0.043 ms} & 669 MB & 436 sec \\ 
			& Faiss (IVFADC)~\cite{web_faiss, tpami_jegou2011} & $K=10^3, M=64, n_{\mathrm{probe}}=4$ & 0.67 & 0.61 ms & 73 MB & 30 sec \\ 
			& Rii (proposed) & $K=10^3, M=64, L=5000$ & 0.64 & 0.73 ms & \textbf{69 MB} & 82 sec \\ 
			& Rii-OPQ (proposed) & $K=10^3, M=64, L=5000$ & 0.65 & 0.82 ms & \textbf{69 MB} & 85 sec \\  \midrule
			\multirow{6}{*}{GIST1M}	& Annoy~\cite{web_annoy} & $n_{\mathrm{trees}}=2000$, $k_{\mathrm{search}}=2000$ & 0.49 & 1.2 ms & 5023 MB & 2088 sec  \\ 
			& FALCONN~\cite{web_falconn, nips_andoni2015} & $n_{\mathrm{probes}}=512$ & 0.53 & 8.6 ms & - & \textbf{7.2 sec} \\ 
			& NMSLIB (HNSW)~\cite{web_nmslib, sisap_boytsov2013, corr_malkov2016} & $\mathrm{efS}=8$ & 0.49 & \textbf{0.19 ms} & 3997 MB & 1576 sec \\ 
			& Faiss (IVFADC)~\cite{web_faiss, tpami_jegou2011} & $K=10^3, M=240, n_{\mathrm{probe}}=8$ & 0.52 & 3.8 ms & 253 MB & 51 sec \\ 
			& Rii (proposed) & $K=10^3, M=240, L=8000$ & 0.45 & 3.2 ms & \textbf{246 MB} & 353 sec \\ 
			& Rii-OPQ (proposed) & $K=10^3, M=240, L=8000$ & 0.50 & 3.8 ms & 249 MB & 388 sec \\  
			\bottomrule
		\end{tabular}
	\end{center}
	\vspace{-2mm}		
\end{table*}

Table~\ref{tbl:million_compare} presents the results.
We summarize our findings:
\begin{itemize} [leftmargin=\ItemizeLeftMargin]
	\item Rii was comparable with the state-of-the-art system Faiss.
	In particular, although our method is basically an approximation of IVFADC,
	the decrease in the accuracy is not significant.
	\item Rii was the most memory efficient among the methods. 
	The measured value is almost same as the theoretically predicted value (68 MB against 69 MB and 244 MB against 249 MB).	
	\item If we compare Rii and Rii-OPQ, Rii-OPQ was slightly slower but a little more accurate with the same parameter settings.
	\item Annoy achieved the second fastest result. Because Annoy supports the direct memory map system,
	the construction required some time and consumed a relatively large disk space.
	\item FALCONN achieved a comparable (or slightly slower) performance to Faiss/Rii.
	We note that the building cost of FALCONN is considerably smaller than for other methods.
	As FALCONN does not provide IO functions, we did not report the disk space.
	\item As reported in the benchmark~\cite{sisap_aumuller2017, web_annbenchmark},
	NMSLIB achieved the fastest performance.
	On the other hand, the building time and memory consumption are inferior relative to Faiss/Rii.
	\item The results for SIFT1M and GIST1M follow similar tendencies.
\end{itemize}

\section{Application}
We present an application to highlight the subset search function of Rii.
For this demonstration, we leverage the data of 
The Metropolitan Museum of Art (MET) Open Access\footnote{\url{https://github.com/metmuseum/openaccess}}.
This dataset contains more than 420,000 items
from MET, with both the image and extensive metadata for each item (\Tref{tbl:metadata}).
From this data, we select 201,998 items that are provided with the Creative Common license.
For each image, we extracted a 1,920-dimensional
activation of last average pooling layer of the DenseNet-201~\cite{cvpr_huang2017b} architecture trained with ImageNet.
The features are stored in Rii with $M=192$. 
Several meta-information is stored in a table using Pandas\footnote{\url{https://pandas.pydata.org/}},
which is a popular on-memory data management system for Python.

\begin{figure}
	\begin{minted}[mathescape, % 数式使える
	%linenos, % 行番号
	%fontfamily=courier, % いい感じのフォント
	breaklines, % 長すぎる行をうまく改行
	breakindent=20pt, % 改行後の左からの幅
	fontsize=\small, % フォントサイズ
	%	numbersep=1pt, % 行番号の内側よせ余白？
	frame=single, % lines:上下に線 single:線で囲う
	%baselinestretch=0.7, % 行間
	]{python}
import pandas as pd
import rii 

# Read data
df = pd.read_csv('metadata.csv')
engine = pkl.load(open('rii_densenet.pkl', 'rb'))

# Metadata search (13.5 ms)
S = df[(df['data']<500) & (df['country']=='Egypt')]['ID']
S = np.sort(np.array(S))  # Target identifiers

# ANN for subset (2 ms)
q = # Read query feature
result = engine.query(q=q, target_ids=S, topk=3)
	\end{minted}

	\begin{tabular}{@{}ccccc@{}}
		\includegraphics[height=0.26\linewidth]{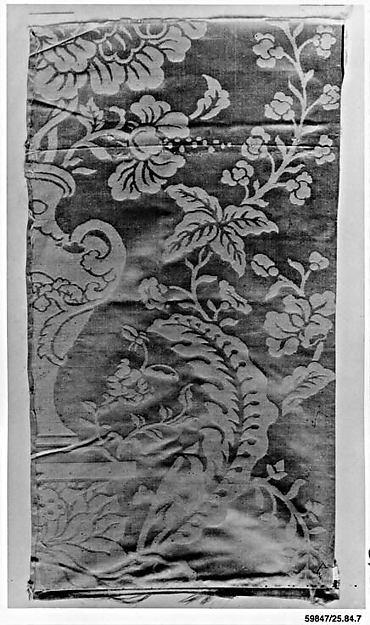} &
		\includegraphics[height=0.26\linewidth]{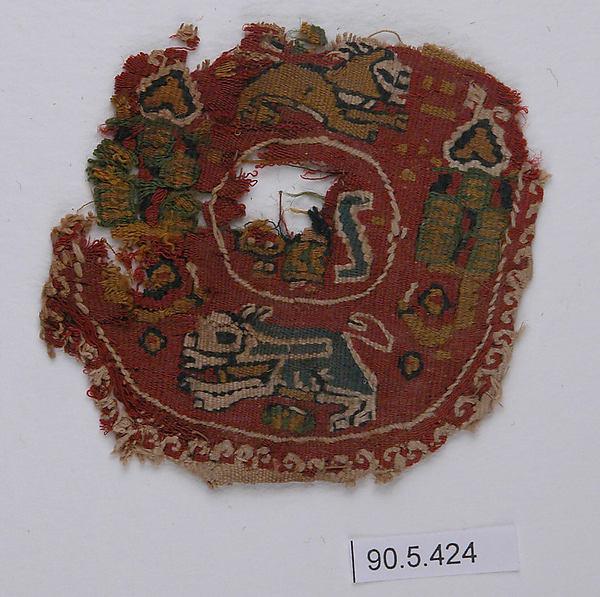} &
		\includegraphics[height=0.26\linewidth]{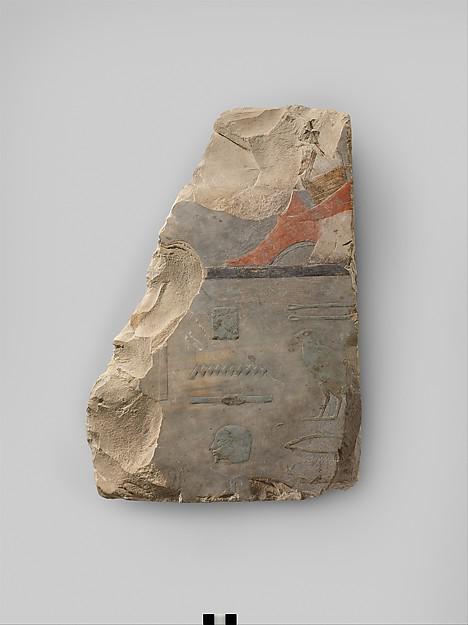} &
		\includegraphics[height=0.26\linewidth]{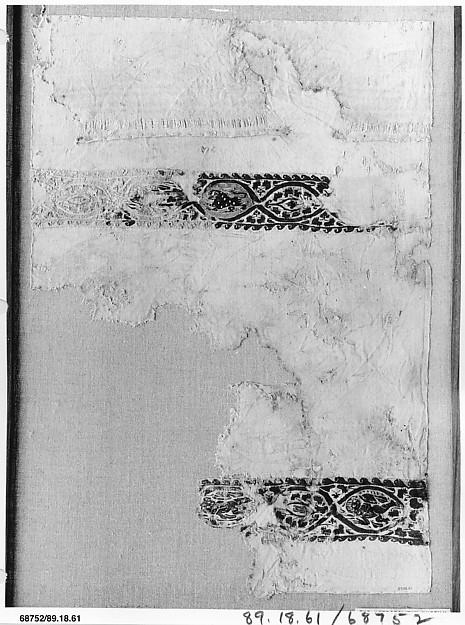} \\
		\footnotesize{Query} & \footnotesize{The nearest} & \footnotesize{The 2nd nearest} & \footnotesize{The 3rd nearest}
	\end{tabular}	
	
	\caption{
		Demonstration of the subset search. The target items are first selected using metadata information.
		Then, an image-based search is conducted over the target items.
	}
	\label{fig:application}
\end{figure}

\Fref{fig:application} demonstrates the system,
including Python codes and the search results.
The metadata and DenseNet vectors are first read.
Then, the search is conducted based on the metadata.
Here, the items that were created before A.D. 500 in Egypt are specified.
Next, the target identifiers $\mathcal{S}$ are prepared. This is simply a set of IDs of the 
selected items.
The image-based search is then conducted over them.
The query here is Chinese tapestry.
We can find similar items to the Chinese tapestry from the museum items in ancient Egypt.

As this demonstration reveals, the search using the target subset is a general problem setting.
Rii can solve this type of problem easily.
As \Sref{sec:exp_subset} shows, existing methods using the late checking module do not perform well
when $|\mathcal{S}|$ is small.
For example, in this case the result of the metadata search can have any number of items. 
Rii can handle a subset search for any size of $\mathcal{S}$.

\begin{table}
	\begin{center}
		\caption{
			Metadata of MET dataset. Each item has several attributes, such as title and data.
		}
		\vspace{-2mm}
		\label{tbl:metadata}				
		\begin{tabular}{@{}lllll@{}} \toprule %
			ID & title & date & country & $\cdots$ \\ \midrule
			0 & Bust of Abraham Lincoln & 1876 & United States & \\
			1 & Acorn Clock & 1847 & United States & \\
			& $\vdots$ & & & \\
			\bottomrule
		\end{tabular}
	\end{center}
\end{table}

\section{Conclusions}
We developed an approximate nearest neighbor search method, called Rii.
Rii provides the two functions of searching over a subset and a reconfigure function
for newly added vectors.
Extensive comparisons showed that Rii achieved a comparable
performance to state-of-the art systems, such as Faiss.

Note that the latest systems incorporate HNSW for the coarse assignment of IVFADC~\cite{eccv_baranchuk2018, cvpr_douze2018}.
Our Rii architecture can be combined to them, but that will be remained as a future work.

\textbf{Acknowledgments:}
This work was supported by JST ACT-I Grant Number JPMJPR16UO, Japan.

\bibliographystyle{ACM-Reference-Format}
\bibliography{sigproc}

\end{document}